\documentclass{article}

\PassOptionsToPackage{numbers, compress}{natbib}

\usepackage[preprint]{nips_2018}

\usepackage[utf8]{inputenc} %
\usepackage[T1]{fontenc}    %
\usepackage{booktabs}       %
\usepackage{amsfonts}       %
\usepackage{amssymb}
\usepackage{amsmath}
\usepackage{color}
\usepackage{graphicx}
\usepackage{nicefrac}       %
\usepackage{microtype}      %
\usepackage{caption}
\usepackage{subcaption}
\usepackage{algorithm}
\usepackage[noend]{algpseudocode}
	\makeatletter
	\def\BState{\State\hskip-\ALG@thistlm}
	\makeatother
\usepackage{tikz}
\newcommand{\G}[3]{\mathcal N \left( {#1} | {#2}, {#3} \right)}

\newcommand{\KL}[2]{{\rm KL} \left({#1} \parallel {#2} \right)}
\newcommand{\mc}{\mathcal}
\newcommand{\mtx}{\mathbf}
\newcommand{\order}[1]{\mathcal O \left( {#1} \right)}
\newcommand{\reals}{\mathbb R}
\newcommand{\vc}{\boldsymbol}

\newcommand{\determinant}[1]{\left| {#1} \right|}

\newcommand{\diag}[0]{\textrm{diag}}

\newcommand{\expectation}[2]{\mathbb E_{#2} \left[ {#1} \right]}

\newcommand{\trace}[1]{\textrm{Tr}\left( {#1} \right)}

\newcommand{\xxi}[0]{\mtx X^{(\xi)}}
\newcommand{\xxis}[0]{\mtx X^{(\xi, *)}}

\newcommand{\xsts}[0]{\mtx X^{(s, *)}}
\newcommand{\xu}[0]{\mtx X_u}
\newcommand{\xuxi}[0]{\mtx X_u^{(\xi)}}
\newcommand{\xust}[0]{\mtx X_u^{(s)}}

\newcommand{\Ubar}[0]{\bar{\mtx U}}
\newcommand{\ubar}[0]{\bar{\vc u}}
\newcommand{\kff}[0]{\mtx K_{ff}}

\newcommand{\kfu}[0]{\mtx K_{fu}}
\newcommand{\kuf}[0]{\mtx K_{uf}}
\newcommand{\kuu}[0]{\mtx K_{uu}}
\newcommand{\kuuinv}[0]{\kuu^{-1}}
\newcommand{\ksu}[0]{\mtx K_{*u}}
\newcommand{\kus}[0]{\mtx K_{u*}}
\newcommand{\kss}[0]{\mtx K_{**}}
\newcommand{\ktilde}[0]{\tilde{\mtx K}}

\newcommand{\psione}[0]{\mtx \Psi_1}
\newcommand{\psitwo}[0]{\mtx \Psi_2}
\newcommand{\kpsi}[0]{\mtx K_\psi}
\newcommand{\kpsiinv}[0]{\kpsi^{-1}}

\title{Structured Bayesian Gaussian process latent variable model}

\author{
    Steven Atkinson %
    \\
    Center for Informatics and\\
    Computational Science\\
    University of Notre Dame\\
    \texttt{satkinso@nd.edu}
    \And
    Nicholas Zabaras\thanks{\url{zabaras.com}} \\
    Center for Informatics and\\
    Computational Science\\
    University of Notre Dame\\
    \texttt{nzabaras@nd.edu} \\
}

\begin{document}

\maketitle

\begin{abstract}
    We introduce a Bayesian Gaussian process latent variable model that explicitly captures spatial correlations in data using a parameterized spatial kernel and leveraging structure-exploiting algebra on the model covariance matrices for computational tractability.
    Inference is made tractable through a collapsed variational bound with similar computational complexity to that of the traditional Bayesian GP-LVM.
    Inference over partially-observed test cases is achieved by optimizing a ``partially-collapsed'' bound.
    Modeling high-dimensional time series systems is enabled through use of a dynamical GP latent variable prior.
    Examples imputing missing data on images and super-resolution imputation of missing video frames demonstrate the model.
\end{abstract}

\section{Introduction}
\label{sec:Intro}

Probabilistic generative models are valuable for inferring structure in large sets of unlabeled data~\cite{titsias2010bayesian}.
Bayesian generative models such as the Gaussian process latent variable model~\cite{lawrence2004gaussian, lawrence2005probabilistic, titsias2010bayesian} and the related unsupervised deep Gaussian processes~\cite{damianou2013deep, dai2015variational} leverage the expressive, yet regularized flexibility of Gaussian processes within an unsupervised learning setting, producing low-dimensional nonlinear embeddings of data. 
Analytical tractability is achieved through the variational inference procedure first given by Titsias~\cite{titsias2009variational} and later extended by Titasis and Lawrence~\cite{titsias2010bayesian} for the case of unobserved inputs.
The dynamical GP-LVM~\cite{damianou2016variational} constrains the learned latent variables through a Gaussian process prior parameterized by time points corresponding to the observed examples.

However, the generative models utilized by all of the above mentioned do not account for the possibility that the data being modeled exhibit correlations.
From a modeling standpoint, the selection of a prior that factorizes across dimensions is rather uninformative and adversely affects model sample efficiency.
Instead, one might wish to encode knowledge about correlations within each high-dimensional observation through the model prior.
Beyond considerations of sample efficiency, because no spatial information is incorporated explicitly into the model, one is not able to use these approaches for inference at spatial locations not included in the training set.

At the same time, neural networks have made great progress as generative models in recent years via variational autoencoder architectures~\cite{kingma2013auto} and adversarial formulations~\cite{goodfellow2014generative}. 
A main focus of this work is sample efficiency within unsupervised learning;
as of now, the above generally struggle to perform well with a limited supply of data.
A reasonable means of addressing this challenge is by carefully-chosen Bayesian regularization;
however, we are still challenged to pose tractable, interpretable priors and methods for inference.
While imposing Gaussian or Laplace priors over the model weights \cite{blundell2015weight} is attractive because of its mathematical tractability~\cite{goodfellow2016deep}, one may still question 
how well one understands the distributions over functions they imply.
to what extent one can understand the implications of such priors on the resulting distribution over functions that the model inherits~\cite{salimbeni2017doubly}.
Indeed, merely \textit{interpreting} workable neural network priors is a challenging task and is of considerable interest in its own right.
In this work, our approach begins by defining the generative model's distribution over functions and proceeding to derive a tractable means of inference.

The main contribution of this work is to extend the Bayesian GP-LVM by utilizing structure-exploiting algebra~\cite{saatcci2012scalable} to enable efficient inference of latent variables and a generative model with explicit spatial correlations modeled by a parameterized kernel function.
We call this the structured Gaussian process latent variable model (SGPLVM).
The learned generative model may be used for data imputation for partially-observed test data.
Furthermore, the dynamical prior of Damianou et al.~\cite{damianou2016variational} is also incorporated into the model to allow for interpolation for high-dimensional time series data.
For tractable inference, we derive a collapsed variational bound that may be efficiently computed by exploiting Kronecker product structure, achieving the same computational complexity as a typical sparse Gaussian process model.

The model bears a resemblance to the linear model of coregionalization~\cite{journel1978mining, goovaerts1997geostatistics}.
However, our model extends this by allowing the inputs to be uncertain within the usual variational free energy approach first shown in~\cite{titsias2010bayesian}.
Computational tractability is maintained by exploiting the structure of the covariance matrix by extending the structured GP methods first introduced by Saat\c{c}i~\cite{saatcci2012scalable}.
Additionally, the modeled spatial correlations are expressed in terms of a familiar kernel function (see, e.g., \cite{rasmussen2006gaussian}) with a simple parameterization that is simple to optimize over, given high-dimensional observation data, and interpretable through its hyperparameters.

The rest of this paper is organized as follows.
In Section~\ref{sec:Theory}, we define the structured GP-LVM including its variational lower bound, predictive density, and algorithm for inference of latent variables.
In Section~\ref{sec:Examples}, we apply our model to model high-dimensional images with and without a dynamical prior and demonstrate its ability to impute missing data in several novel ways.
Section~\ref{sec:Conclusion} summarizes our work and discusses its implications.

\subsection{Related work}
Our model is based on the Bayesian Gaussian process latent variable model of Titsias and Lawrence~\cite{titsias2010bayesian}.
Structured Gaussian process regression was first introduced in~\cite{saatcci2012scalable}, who considered inputs composed as a Cartesian product of one-dimensional inputs, though combinations of higher-dimensional inputs are also possible so long as the kernel is appropriately separable \cite{bilionis2013multi}.
The insight to exploit Kronecker product structure in Gaussian process latent variable models was first shown in~\cite{lopez2017efficient}.
However, their variational formulation defines a posterior over the induced outputs $\mtx U$ and restricts its covariance matrix to possess Kronecker product structure;
The $\order{m}$ variational parameters describing its mean and covariance must be learned through optimization (i.e.\ the bound is ``uncollapsed'' \cite{damianou2016variational}).
Here, we make no assumptions about the structure of the posterior covariance, instead deriving the analytical optimum and showing how the resulting ``collapsed'' variational bound can still be computed efficiently by exploiting eigendecomposition properties of Kronecker products.
Inference over partially-observed data is still possible using a novel `mixed'' bound combining collapsed and uncollapsed terms from the training and test data, respectively.
We also make the connection with the dynamical GP-LVM of~\cite{damianou2016variational} for modeling temporally-correlated data.

\section{Theory}
\label{sec:Theory}

In this section, we derive the structured GP-LVM (SGP-LVM) model.
Before proceeding, we quickly define some useful notation.
Let $\mtx X$ be a matrix with $i$-th row $\vc x_{i, :}$, $j$-th column $\vc x_{:, j}$, and entry $x_{ij}$.
The matrix $\mtx X$ may be assembled by a set of vectors $\vc x$ as rows of $\mtx X$.

\subsection{Problem statement}
\label{sec:Theory:ProblemStatement}
Suppose we are given a set of $n_\xi$ data, each of which is observed simultaneously with $d_y$ channels at $n_s$ points in a $d_s$-dimensional spatial domain $\mc X_s \subset \reals^{d_s}$.
Represent this data as a matrix $\hat{\mtx Y} \in \reals^{n_\xi \times n_s d_y}$.
The observations may also be dynamical in nature and indexed with time points $\vc t \in \reals^{n_\xi}$.
The spatial inputs are assembled as a matrix $\mtx X^{(s)} \in \reals^{n_s \times d_s}$.
Alternatively, instead of modeling our observations as $n_\xi$ data that are $n_s d_y$-dimensional (``few data, many dimensions''), we may instead reshape $\hat{\mtx Y}$ to model them as
$n = n_\xi n_s$ data that are $d_y$-dimensional (``many data, few dimensions'').
Suppose that each of the $n_\xi$ data are described by a latent variable $\vc x^{(\vc \xi)} \in \mc X_\xi \subset \reals^{d_\xi}$; 
represent these as the matrix $\mtx X^{(\xi)} \in \reals^{n_\xi \times d_\xi}$ with a prior $p(\mtx X^{(\xi)})$.
The $d_x = d_\xi + d_s$-dimensional inputs are defined as the Cartesian product of the spatial and latent variable inputs: $\mtx X = \mtx X^{(\xi)} \times \mtx X^{(s)} \in \reals^{n \times d_x}$, with $n = n_\xi n_s$.\footnote{One can further impose structure within the spatial inputs, i.e., $\mtx X^{(s)} = \mtx X^{(s, 1)} \times \dots \times \mtx X^{(s, d_s)}$.}
We aim to learn a generative model
\begin{equation}
    \vc y = \vc f(\vc x) + \vc \epsilon, ~\vc f: \mc X_s \times \mc X_\xi \rightarrow \reals^{d_y} \sim \mc{GP}, \vc \epsilon \sim \G{\epsilon}{\vc 0}{\sigma_y^2 \mtx I_{d_y \times d_y}},
    \label{eqn:ProblemStatement:GenerativeModel}
\end{equation}
The model definition and training procedure is described in Sec.~\ref{sec:Theory:SGPLVM:Training}.

Given the learned model, we seek to carry out two types of predictions.
First, given a test example $\mtx Y^* \in \reals^{n_s^* \times d_y}$ observed at $n_s^*$ spatial points $\mtx X^{(s, *)} \in \reals^{n_s^* \times d_s}$, we seek to infer the posterior of its latent variable, $q(\vc x^{(\xi, *)})$.
Second, given $q(\vc x^{(\xi, *)})$ and some (possibly different) test points in space $\mtx X^{(s, *)}$, we wish to compute the predictive density for the corresponding outputs $\mtx Y^* \in \reals^{n_s^* \times d_y}$.
These are addressed in Sec.~\ref{sec:Theory:SGPLVM:Predictions}.

\subsection{Structured Bayesian Gaussian process latent variable model}
\label{sec:Theory:SGPLVM}

In the following subsections, we explain how to train and do predictions with the structured Gaussian process latent variable model.

\subsubsection{SGPLVM: model architecture and evidence lower bound}
\label{sec:Theory:SGPLVM:Training}

As mentioned in~\cite{lopez2017efficient}, inference for the generative model of Eq.~(\ref{eqn:ProblemStatement:GenerativeModel}) is challenging for two reasons.
First, computing the marginal likelihood $p(\mtx Y | \mtx X^{(s)}) = \int p(\mtx Y | \mtx X) p(\mtx X^{(\xi)}) d \mtx X^{(\xi)}$ is analytically intractable.
Second, the likelihood $p(\mtx Y | \mtx X)$ requires the inversion of a $n \times n$ covariance matrix.
Using the approach of Titsias and Lawrence, the model is augmented with $m$ inducing input-output pairs, assembled in matrices $\mtx X_u \in \reals^{m \times d_x}$ and $\mtx U \in \reals^{m \times d_y}$;
we assume that these inducing pairs are modeled by the same generative process as the training inputs and latent outputs $\mtx F \in \reals^{n \times d_y}$, allowing us to write the following joint probability:
\begin{equation}
    p(\mtx F^+ | \mtx X^+) = \prod_{j=1}^{d_y} \G{\vc f_{:, j}^+}{\vc m_{:, j}^+}
    {\left( \begin{matrix} \kff & \kfu \\ \kuf & \kuu \end{matrix} \right)},
    ~
    \mtx X^+ = \left( \begin{matrix} \mtx X \\ \mtx X_u \end{matrix} \right),
    ~
    \mtx F^+ = \left( \begin{matrix} \mtx F \\ \mtx U \end{matrix} \right).
\end{equation}
where $\kff$, $\kuf$, $\kfu$, and $\kuu$ are kernel matrices evaluated on the training and inducing inputs.
The conditional GP prior is given by
\begin{equation}
    p(\mtx F | \mtx X, \mtx U, \mtx X_u) = \prod_{j=1}^{d_y} \G{\vc f_{:, j}}{\vc \eta_{:, j}}{\ktilde},
\end{equation}
where the conditional mean and covariance take the usual forms from the projected process model~\cite{seeger2003fast, rasmussen2006gaussian}:
\begin{align}
    \vc \eta &= \kfu \kuuinv \mtx Y,
    \\
    \ktilde &= \kff - \kfu \kuuinv \kuf.
\end{align}
Finally, we define the Gaussian variational posterior over the induced outputs
\begin{equation}
    q(\mtx U) = \prod_{j=1}^{d_y} \G{\vc u_{:, j}}{\ubar_{:, j}}{\mtx \Sigma_u},
\end{equation}
and pick a joint variational posterior that factorizes as
\begin{equation}
    q(\mtx F, \mtx U, \xxi) = p(\mtx F | \mtx X, \mtx U, \mtx X_u) q(\mtx U) q(\xxi).
\end{equation}
The form of $q(\xxi)$ will be discussed later.

Using the familiar variational approach~\cite{bishop2006pattern, titsias2010bayesian}, we use Jensen's inequality to write a lower bound for the logarithm of the model evidence. 
Following the usual approach, we use Jensen's inequality to define the following lower bound on the logarithm of the model evidence:
\begin{equation}
    \log p(\mtx Y) \ge \mc L = \int q(\mtx F, \mtx U, \xxi) \log \frac{p(\mtx F, \mtx U, \xxi)}{q(\mtx F, \mtx U, \xxi)} d \mtx F d \mtx U d \xxi.
\end{equation}
After some manipulations, one arrives at the following uncollapsed lower bound:
\begin{equation}
    \begin{aligned}
        \mc L = &-\frac{n d_y}{2} \left( \log(2 \pi) - \log \beta \right)
        -\frac{\beta}{2} \trace{\mtx Y \mtx Y^\intercal}
        \\
        &+ \beta \trace{\Ubar^\intercal \kuuinv \psione^\intercal \mtx Y}
        - \frac{\beta}{2} \trace{\kuuinv \psitwo \kuuinv \left( \Ubar \Ubar^\intercal + d_y \mtx \Sigma_u \right)}
        \\
        &- \KL{q(\mtx U)}{p(\mtx U)} - \KL{q(\xxi)}{p(\xxi)},
    \end{aligned}
    \label{eqn:Theory:SGPLVM:UncollapsedBound}
\end{equation}
where we have defined the kernel expectations
\begin{equation}
    \begin{aligned}
        \psione &= \expectation{\kfu}{q(\xxi)},
        \\
        \psitwo &= \expectation{\kuf \kfu}{q(\xxi)}.
    \end{aligned}
    \label{eqn:Theory:SGPLVM:PsiStatistics}
\end{equation}
Using standard methods \cite{bishop2006pattern}, one may find an analytic optimum for $q(\mtx U)$ that maximizes Eq.\ (\ref{eqn:Theory:SGPLVM:UncollapsedBound}) with respect to $\Ubar$ and $\mtx \Sigma_u$:
\begin{equation}
    q^*(\mtx U) = \prod_{j=1}^{d_y} \G{\vc u_{:, j}}{\ubar_{:, j}^*}{\mtx \Sigma_u^*},
    ~
    \Ubar^* = \kuu \kpsiinv \psione^\intercal \mtx Y,
    ~
    \mtx \Sigma_u^* = \beta^{-1} \kuu \kpsiinv \kuu,
    \label{eqn:Theory:SGPLVM:OptimalU}
\end{equation}
where we have defined
\begin{equation}
    \kpsi =  \beta^{-1} \kuu + \psitwo
    \label{eqn:Theory:SGPLVM:KPsi}
\end{equation}
for convenience.
Substituting this into Eq.~(\ref{eqn:Theory:SGPLVM:UncollapsedBound}) and doing the required manipulations results in the ``collapsed'' lower bound:
\begin{equation}
    \begin{aligned}
        \mc L = &\frac{d_y}{2} \left(
            (n - m) \log \beta - n \log (2 \pi) - \log \determinant{\mtx A}
        \right)
        \\
        &- \frac{\beta}{2} \left(
            \trace{\mtx Y \mtx Y^\intercal}
            - \trace{\mtx Y^\intercal \mtx \Psi_1 \mtx K_\psi^{-1} \mtx \Psi_1^\intercal \mtx Y}
            + d_y \left(
                \psi_0 - \trace{\mtx C}
            \right)
        \right)
        \\
        &- \KL{q(\xxi)}{p(\xxi)},
    \end{aligned}
    \label{eqn:Theory:SGPLVM:CollapsedBound}
\end{equation}
where 
\begin{align}
    \psi_0 &= \expectation{\trace{\kff}}{q(\xxi)},
    \\
    \mtx C &= \mtx L^{-1} \psitwo \mtx L^{-\intercal},
    \\
    \mtx A &= \mtx L^{-1} \kpsi \mtx L^{-\intercal} = \beta^{-1} \mtx I_{m \times m} + \mtx C,
\end{align}
and 
$\kuu = \mtx L \mtx L^\intercal$ 
is a Cholesky decomposition.
For modeling independent, identically-distributed data, we use the prior $p(\xxi) = \prod_{ij} \G{x_{ij}^{(\xi)}}{0}{1}$ and variational posterior $q(\xxi) = \prod_{ij} \G{x_{ij}^{(\xi)}}{\mu_{ij}^{(\xi)}}{c_{ij}^{(\xi)}}$.
For dynamical data indexed by times $\{ t_i \}_{i=1}^{n_\xi}$, we follow~\cite{opper2009variational, damianou2016variational} and use the GP prior 
$p(\xxi) = \prod_{j=1}^{d_\xi} \G{\vc x_{:, j}^{(\xi)}}{\vc 0}{\mtx K_{xx}}$ 
and variational posterior 
$q(\xxi) = \prod_{j=1}^{d_\xi} \G{\vc x_{:, j}^{(\xi)}}{\mtx K_{xx} \vc \mu_{:, j}^{(\xi)}}{(\mtx K_{xx}^{-1} + \mtx \Lambda^{(x, j)})^{-1}}$,
where $\mtx K_{xx}$ is the covariance matrix formed on $\vc t$ by a kernel $k_t(t, t'; \vc \theta_t)$ and $\mtx \Lambda^{(x, j)}$ is a diagonal $n_\xi \times n_\xi$ matrix.

To train the model, we maximize Eq.~(\ref{eqn:Theory:SGPLVM:CollapsedBound}) over the variational parameters of $q(\xxi)$ the inducing inputs, and the model hyperparameters $\vc \theta = \{\vc \theta_k, \beta\}$.
However, doing so na\"ively quickly becomes computationally intractable.
Next, we show how the structure in the model inputs can be exploited so that the time and memory requirements associated with training become linear in $n$.

Similarly to \cite{lopez2017efficient}, we impose structure on the inducing points:  
$\xu = \xuxi \times \xust$,
where 
$\xuxi \in \reals^{m_\xi \times d_\xi}$ and 
$\xust \in \reals^{m_s \times d_s}$.
This implies that $\kuu = \kuu^{(\xi)} \otimes \kuu^{(s)}$ and $\kfu = \kfu^{(\xi)} \otimes \kfu^{(s)}$.

The statistics of Eq.~(\ref{eqn:Theory:SGPLVM:PsiStatistics}) can then be written as
\begin{equation}
    \psi_0 = \psi_0^{(\xi)} \trace{\kff^{(s)}},
    ~
    \psione = \psione^{(\xi)} \otimes \kfu^{(s)},
    ~
    \psitwo = \psitwo^{(\xi)} \otimes (\kuf^{(s)} \kfu^{(s)}),
    \label{eqn:Theory:SGPLVM:PsiStatisticsStructured}
\end{equation}
where $\psi_0^{(\xi)}$, $\psione^{(\xi)}$, and $\psitwo^{(\xi)}$ may be evaluated in the usual way (see~\cite{damianou2015deep}).
The computational cost associated with evaluating the kernel expectations in Eq.~(\ref{eqn:Theory:SGPLVM:PsiStatisticsStructured}) remains unchanged relative to the traditional Bayesian GP-LVM and still allows for parallelization with respect to latent variables of $\xxi$ as originally discussed in~\cite{dai2014gaussian, gal2014distributed}.

Finally, we show how the remaining terms of the variational bound may be computed efficiently without having to ever explicitly evaluate any of the full $m \times m$ matrices in the bound.
First, we note that $\mtx L$ and $\mtx C$ have Kronecker decompositions:
\begin{align}
    \mtx L &= \mtx L^{(\xi)} \otimes \mtx L^{(s)},
    \\
    \mtx C &= \mtx C^{(\xi)} \otimes \mtx C^{(s)}.
\end{align}
The eigendecomposition of $\mtx C$ is
\begin{equation}
    \mtx C = \mtx Q_C \mtx \Lambda_C \mtx Q_C^\intercal,
\end{equation}
so $\mtx Q_C$ is orthogonal and $\mtx \Lambda_C$ is diagonal.
These both also admit Kronecker decompositions:
\begin{align}
    \mtx Q_C &= \mtx Q_C^{(\xi)} \otimes \mtx Q_C^{(s)},
    \label{eqn:Theory:SGPLVM:QcKronecker}
    \\
    \mtx \Lambda_C &= \mtx \Lambda_C^{(\xi)} \otimes \mtx \Lambda_C^{(s)}.
    \label{eqn:Theory:SGPLVM:LambdaCKronecker}
\end{align}
The matrices on the right hand sides of  Eqs.~(\ref{eqn:Theory:SGPLVM:QcKronecker}) and~(\ref{eqn:Theory:SGPLVM:LambdaCKronecker}) are found by computing the eigendecomposition of $\mtx C^{(\xi)}$ and $\mtx C^{(s)}$.
It follows that
\begin{equation}
    \mtx A = \mtx Q_C \mtx D \mtx Q_C^\intercal.
    \label{eqn:Theory:SGPLVM:AEigen}
\end{equation}
where, for notational convenience, we have defined $\mtx D = \beta^{-1} \mtx I + \mtx \Lambda_C \in \reals^{m \times m}$.
Note that $\mtx D$ is the sum of two diagonals and thus contains $m$ nonzero entries.
Thus, it is at least as easy to store in memory as the data $\mtx Y$.
We also note that expressing $\mtx A$ as in Eq.~(\ref{eqn:Theory:SGPLVM:AEigen}) makes the computation of $\log \determinant{\mtx A}$ in Eq.~(\ref{eqn:Theory:SGPLVM:CollapsedBound}) straightforward.
Next, we see that Eq.~(\ref{eqn:Theory:SGPLVM:KPsi}) can be rewritten as
\begin{equation}
    \kpsi = \mtx L \mtx Q_C \mtx D \mtx Q_C^\intercal \mtx L^\intercal,
\end{equation}
and note that the matrix factors of
\begin{equation}
    \kpsiinv = \mtx L^{-\intercal} \mtx Q_C \mtx D^{-1} \mtx Q_C^\intercal \mtx L^{-1}
    \label{eqn:Theory:SGPLVM:KPsiInverse}
\end{equation}
may be efficiently computed and stored since $\mtx L$ and $\mtx Q_C$ are Kronecker products and the diagonal matrix $\mtx D$ can, of course, be inverted in an element-wise manner.

Finally, the second trace term in the second line of Eq.~(\ref{eqn:Theory:SGPLVM:CollapsedBound}) can be manipulated using Eq.~(\ref{eqn:Theory:SGPLVM:KPsiInverse}) to obtain
\begin{align}
    \trace{\mtx Y^\intercal \mtx \Psi_1 \mtx K_\psi^{-1} \mtx \Psi_1^\intercal \mtx Y} &= \trace{\mtx D^{-1} \underbrace{\mtx Q_C^\intercal \mtx L^{-1} \mtx \Psi_1^\intercal \mtx Y}_{\equiv \mtx B} \mtx Y^\intercal \mtx \Psi_1 \mtx L^{-\intercal} \mtx Q_C}
    \\
    &= \trace{\mtx D^{-1} \mtx B \mtx B^\intercal}
    \\
    &= \sum_{i=1}^m \sum_{j=1}^{d_y} d_{ii}^{-1} b_{ij}^2.
\end{align}
The matrix $\mtx B$ is most efficiently computed by first evaluating the product $\mtx Q_C^\intercal \mtx L^{-1} \mtx \Psi_1^\intercal$, then multiplying against $\mtx Y$ last.
Computing this term takes $\order{n d_y}$ time.
Thus, we see that computing the bound is linear in the size of the training data in both time and memory.

\subsubsection{SGPLVM: predictions}
\label{sec:Theory:SGPLVM:Predictions}

\paragraph{Forward predictive density}
\label{sec:Theory:SGPLVM:Predictions:Forward}
Given test inputs $\mtx X^* = \xxis \times \xsts \in \reals^{n^* \times d_x}$, the predictive density follows form our assumption that the inducing points are sufficient statistics of the training examples~\cite{damianou2016variational}:
\begin{align}
    p(\mtx F^* | \mtx X^*) &= \prod_{j=1}^{d_{out}} \G{\vc f_{:, j}^*}{\vc \mu_{:, j}^*}{\mtx \Sigma^*},
    \label{eqn:Theory:SGPLVM:PredictiveDensity}
    \\
    \vc \mu^* &= \ksu \kpsi^{-1} \mtx \Psi_1^\intercal \mtx Y,
    \label{eqn:Theory:SGPLVM:x_to_y_mean}
    \\
    \mtx \Sigma^* &= \mtx K_{**} 
    - \ksu 
    \left( \kuuinv - \beta^{-1} \kpsi^{-1} \right) 
    \kus,
    \label{eqn:Theory:SGPLVM:x_to_y_covariance}
\end{align}
where $\kss$ is the covariance matrix computed on the test inputs and $\ksu$ is the cross-covariance matrix between test and training inputs.
Noticing that 
$\kss = \kss^{(\xi, *)} \otimes \kss^{(s, *)}$,
$\ksu = \ksu^{(\xi, *)} \otimes \ksu^{(s, *)}$, and substituting Eq.~(\ref{eqn:Theory:SGPLVM:KPsiInverse}) into Eqs.~(\ref{eqn:Theory:SGPLVM:x_to_y_mean}) and~(\ref{eqn:Theory:SGPLVM:x_to_y_covariance}), we obtain
\begin{align}
    \vc \mu^* &= \ksu \mtx L^{-\intercal} \mtx Q_C \mtx D^{-1} \mtx Q_C^\intercal \mtx L^{-1} \mtx \Psi_1^\intercal \mtx Y,
    \label{eqn:Theory:SGPLVM:x_to_y_mean2}
    \\
     \diag (\mtx \Sigma^*) 
    &= 
    \diag(\mtx K_{**}) 
    - \left( \ksu \mtx L^{-\intercal} \mtx Q_C \right) 
    \circ \left( \ksu \mtx L^{-\intercal} \mtx Q_C \right) 
    \diag \left(\mtx I - \beta^{-1} \mtx D^{-1} \right),
\end{align}
where $\circ$ denotes the Schur product.
These may be efficiently computed by exploiting Kronecker products and that $\mtx D$ is diagonal.
Importantly, we are able to use this model to predict at different spatiotemporal resolutions from that of our training data if desired.
This cannot be done without the parameterized kernel that captures spatiotemporal correlations in our model.
The full covariance of Eq.~(\ref{eqn:Theory:SGPLVM:x_to_y_covariance}) can be computed in $\order{m (n_s^{*})^2}$ time and $\order{(n_s^*)^2}$ memory; details are given in the supplementary material.

If we have a Gaussian posterior $q(\xxis)$, then we can readily compute
\begin{equation}
    \bar{\vc \mu}^* = \expectation{\mtx F^*}{q(\xxis)} = \psione^* \kpsi^{-1} \mtx \Psi_1^\intercal \mtx Y,
    \label{eqn:Theory:SGPLVM:prediction:marginal_mean}
\end{equation}
where $\psione^* = \expectation{\ksu}{q(\xxis)} = \psione^{(\xi, *)} \otimes \ksu^{(s)}$.
However, unlike in~\cite{titsias2010bayesian}, the marginal covariance of the predictive density $\bar{\mtx \Sigma}^*$ is intractable due to the modeled spatial correlations.
One can approximate the density as a mixture of Gaussians by taking $n_{MOG}$ samples from $q(\xxis)$ and computing the Gaussian of Eq.~(\ref{eqn:Theory:SGPLVM:PredictiveDensity}) for each sample of $\xxis$.

For the dynamical model, the latent variable posterior at time $t^*$ can be computed as~\cite{damianou2016variational}
\begin{equation}
    q(\vc x^{(\xi, *)}) = \prod_{j=1}^{d_\xi} \G{x_{:, j}^{(\xi, *)}}{\mtx K_{*x} \vc \mu_{:, j}^{(\xi)}}{k_t(t^*, t^*; \vc \theta_t) - \mtx K_{*x} (\mtx K_{xx} + \mtx \Lambda^{(x, j), -1})^{-1} \mtx K_{x*}},
    \label{eqn:Theory:SGPLVM:Prediction:DynamicalPosterior}
\end{equation}
where $\mtx K_{*x}$ is the cross-covariance vector computed on $t^*$ and $\vc t$.

\paragraph{Inference of latent variables}
\label{sec:Theory:SGPLVM:Predictions:Backward}
Here, we explain how the SGPLVM can be used to infer the variational posterior over the latent variables associated with test observations held out from the training set.
Given some test observation $\mtx Y^* \in \reals^{n_s^* \times d_y}$ observed at spatial points $\mtx X^{(s, *)}$, we would like to infer the posterior $q(\vc x^{(\xi, *)})$.
To do this, we optimize the augmented bound
\begin{equation}
    \log p(\mtx Y^*, \mtx Y) \ge \mc L + \mc L^*,
\end{equation}
where 
\begin{equation}
    \begin{aligned}
        \mc L^* = &-\frac{n^* d_y}{2} \left( \log(2 \pi) - \log \beta \right) 
        - \frac{\beta  d_y}{2} \trace{\mtx Y^* \mtx Y^{*, \intercal}}
        - \frac{\beta}{2} \trace{
            \kuu^{-1} \psitwo^* \kuu^{-1} (\Ubar \Ubar^\intercal + \mtx \Sigma_u)
        }
        \\
        &+ \beta \trace{\mtx Y^{*, \intercal} \psione^* \kuu \Ubar}
        - \frac{\beta d_y}{2} \left( \psi_0^* - \trace{\kuu^{-1} \psitwo^*} \right)
        - \KL{q(\vc x^{(\xi, *)})}{p(\vc x^{(\xi, *)})},
    \end{aligned}
    \label{eqn:Theory:SGPLVM:BoundTestTerm}
\end{equation}
and $\Ubar$ and $\mtx \Sigma_u$ are given by Eq.~(\ref{eqn:Theory:SGPLVM:OptimalU}).
Our approach may be seen as a combining the ``collapsed'' bound of~\cite{titsias2010bayesian} to compute $\mc L$ and the ``uncollapsed'' bound used in~\cite{lopez2017efficient} for $\mc L^*$.
Note that, since the test data are not included in $\mc L$, $q(\mtx U)$ is no longer optimal, but the benefit of computational tractability greatly outweighs this drawback in practice.
Equation~(\ref{eqn:Theory:SGPLVM:BoundTestTerm}) may be efficiently computed by exploiting Kronecker structure; details are given in the supplementary material.
To infer $q(\vc x^{(\xi, *)})$, one optimizes $\mc L + \mc L^*$ over its variational parameters.
In the interest of computational efficiency, we keep the variational parameters and kernel hyperparameters learned at training time constant.

\section{Examples}
\label{sec:Examples}

\subsection{Reconstruction of missing data}

In this example, we demonstrate the SGPLVM's ability to reconstruct missing data from an image using the ``Frey Faces'' data set~\cite{roweis2002global,titsias2009variational}, which contains $n_{tot}=1965$ grayscale images at a resolution of $20 \times 28$ pixels and each pixel takes on a scalar value in $[0, 255]$.\footnote{When modeling the data, we rescale the data to have zero mean and unit variance.  However, the results are reported in terms of the raw pixel values.}
We randomly split the data set into $n_\xi$ training and $n_\xi^*$ test images;
we consider the cases with $n_\xi = 50$ and $1000$ and $n_\xi^* = 965$.

As described in Sec.\ \ref{sec:Theory:ProblemStatement}, the training data are assembled into the training data matrix $\mtx Y \in \reals^{n_\xi n_s \times d_y}$, where $n_s = 20 \times 28 = 560$ and $d_y=1$ since the images are grayscale.
The centers of the pixels are assigned two-dimensional spatial inputs and assembled into the spatial input matrix $\mtx X^{(s)} \in \reals^{n_s \times d_s}$ ($d_s = 2$).
The SGPLVM model has $m_\xi=\min(n_\xi, 100)$ latent variable inducing inputs, $m_s = n_s$ spatial inducing inputs, and $d_\xi = 30$ latent dimensions.
Gaussian and Mat\'ern $3/2$ kernels are used for the latent variable and spatial inputs, respectively.

At test time, we randomly remove 50\% of the pixels on each held-out image.
We impute them by inferring the test latent variable posterior following Sec.~\ref{sec:Theory:SGPLVM:Predictions:Backward}, then by approximating the forward predictive density at $\mtx X^{(s, *)} = \mtx X^{(s)}$ as a Gaussian with mean given by Eq.~(\ref{eqn:Theory:SGPLVM:prediction:marginal_mean}) and covariance from a mixture of $n_{MOG}=100$ Gaussians as described in Sec.~\ref{sec:Theory:SGPLVM:Predictions:Forward}.
We condition this Gaussian on the observed pixels in the test observation to repair some of the construction error;
this is only possible because of the spatial correlations modeled by the SGPLVM.
The prediction accuracy is measured for each image by the root mean square error (RMSE) and median negative log probability (MNLP) over the imputed pixels.
We compare our model to the Bayesian GP-LVM on the same data set by replacing the Mat\'ern 3/2 kernel with a white kernel as well as a vanilla across-image GP regression with a Mat\'ern 3/2 kernel trained on each image individually.

\begin{figure}[hbt]
    \centering
    \includegraphics[width=0.7\textwidth]{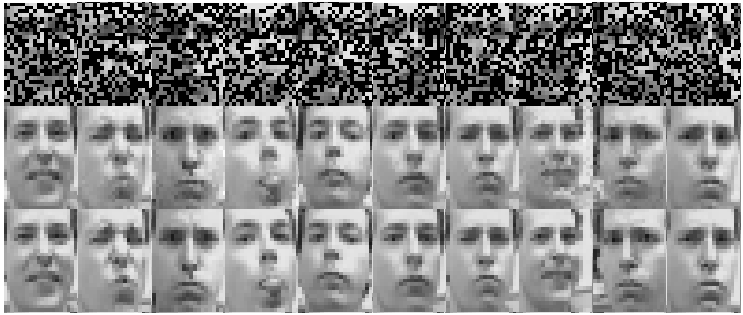}
    \caption{
        (Reconstruction) Example reconstructions using a SGPLVM with $n_\xi=1000$.
        Top row: provided partially-observed realization.
        Middle: Reconstructed observation.
        Bottom: Ground truth.
    }
    \label{fig:Examples:SGPLVM:FreyFacesExamples}
\end{figure}

For $n_\xi = 50$, the SGPLVM attains a mean 
RMSE ($95$th percentiles) of $14.50$ ($6.46$, $28.25$) and
MNLP of $3.25$ ($2.90$, $3.99$);
the Bayesian GP-LVM attains an 
RMSE of $15.63$ ($6.98$, $38.75$) and 
MNLP of $3.45$ ($3.07$, $4.51$).
For $n_\xi=1000$, the SGPLVM attains a mean 
RMSE of $8.91$ ($5.52$, $26.75$) and
MNLP of $3.16$ ($3.02$, $3.38$);
the Bayesian GP-LVM attains an 
RMSE of $9.09$ ($4.39$, $17.47$) and 
MNLP of $3.15$ ($3.00$, $3.53$).
The across-image GP obtains an
RMSE of $19.27$ ($15.06$, $24.30$) and 
MNLP of $6.66$ ($5.92$, $7.92$).
Our results imply that the modeled spatial correlations are helpful particularly when data examples are limited, even though the learned length scales are on the order of a pixel.
However, the across-image GP's performance shows that the latent variable modeling is essential for obtaining good predictions.
This is expected, given the small size of the images.
Figure~\ref{fig:Examples:SGPLVM:FreyFacesExamples} shows a few examples of the reconstructions given by the SGPLVM model.

\subsection{Generation of high-resolution video}
\begin{figure}
    \centering
    \begin{subfigure}[b]{0.3\textwidth}
        \centering
        \includegraphics[width=\textwidth]{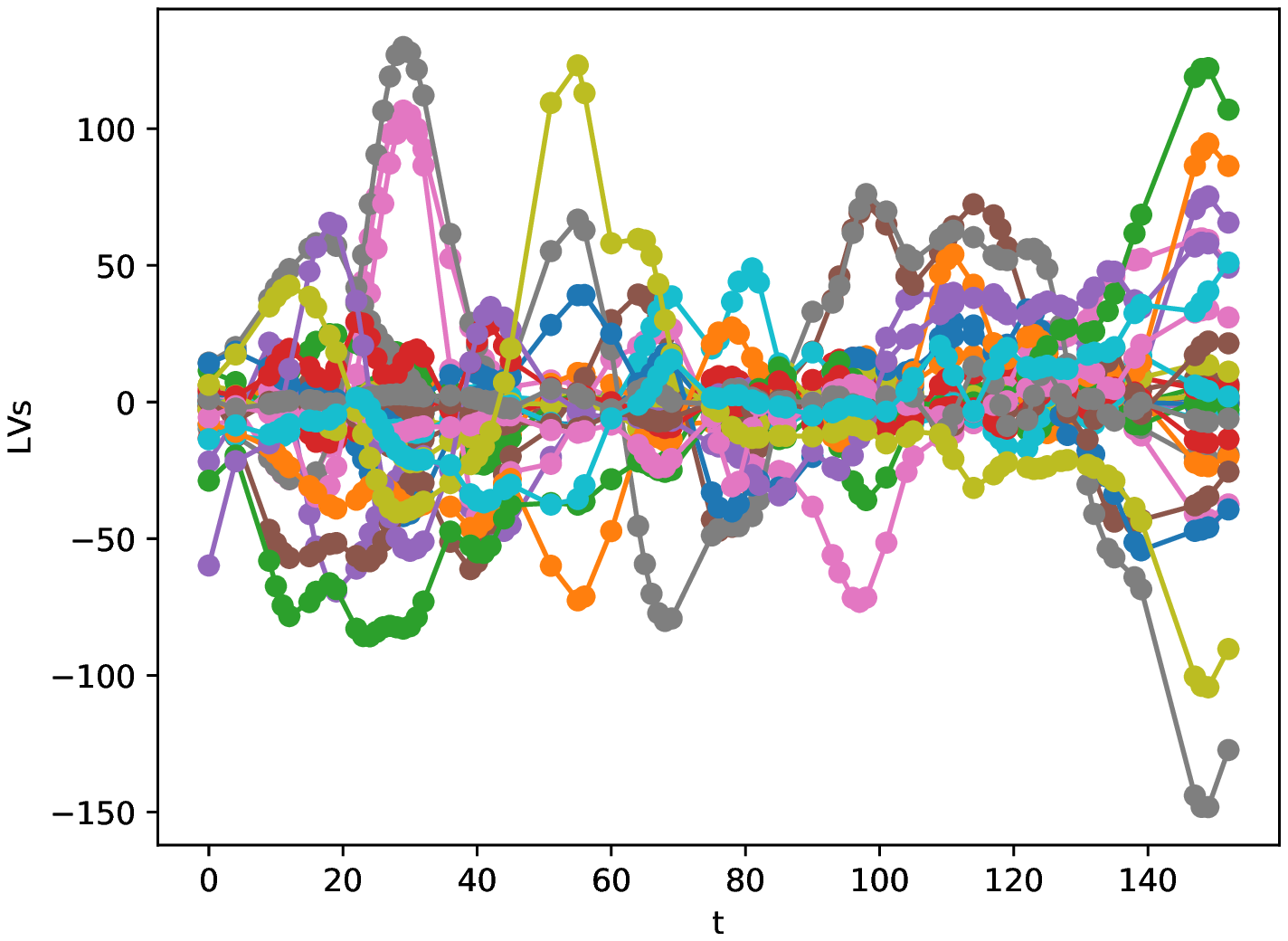}
        \caption{Latent variables}
        \label{fig:Examples:SGPLVM:Video:LVs}
    \end{subfigure}
    ~
    \begin{subfigure}[b]{0.3\textwidth}
        \centering
        \includegraphics[width=\textwidth]{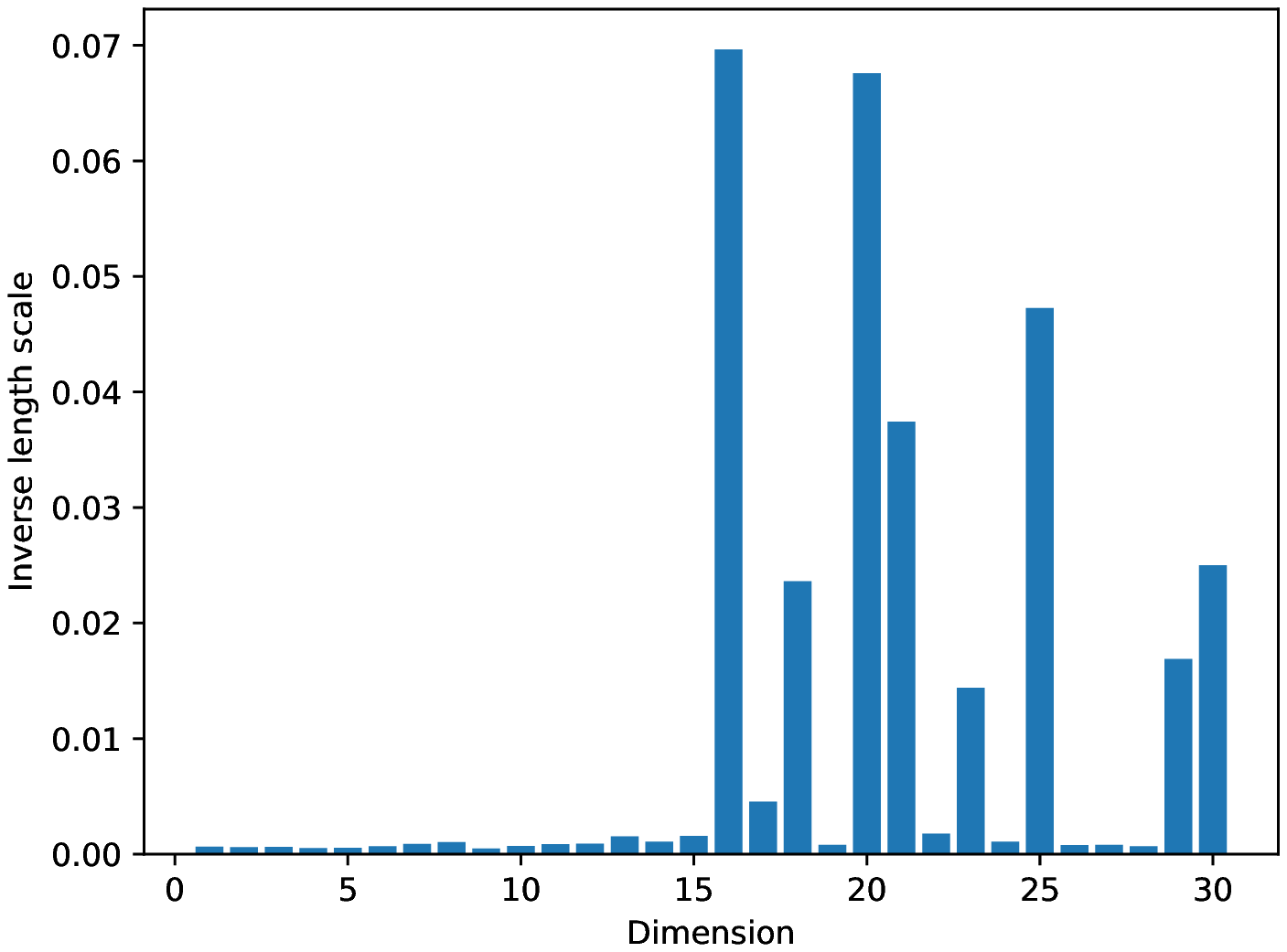}
        \caption{Inverse length scales}
        \label{fig:Examples:SGPLVM:Video:LengthScales}
    \end{subfigure}
    \\
    \begin{subfigure}[b]{0.6\textwidth}
        \centering
        \includegraphics[width=0.45\textwidth]{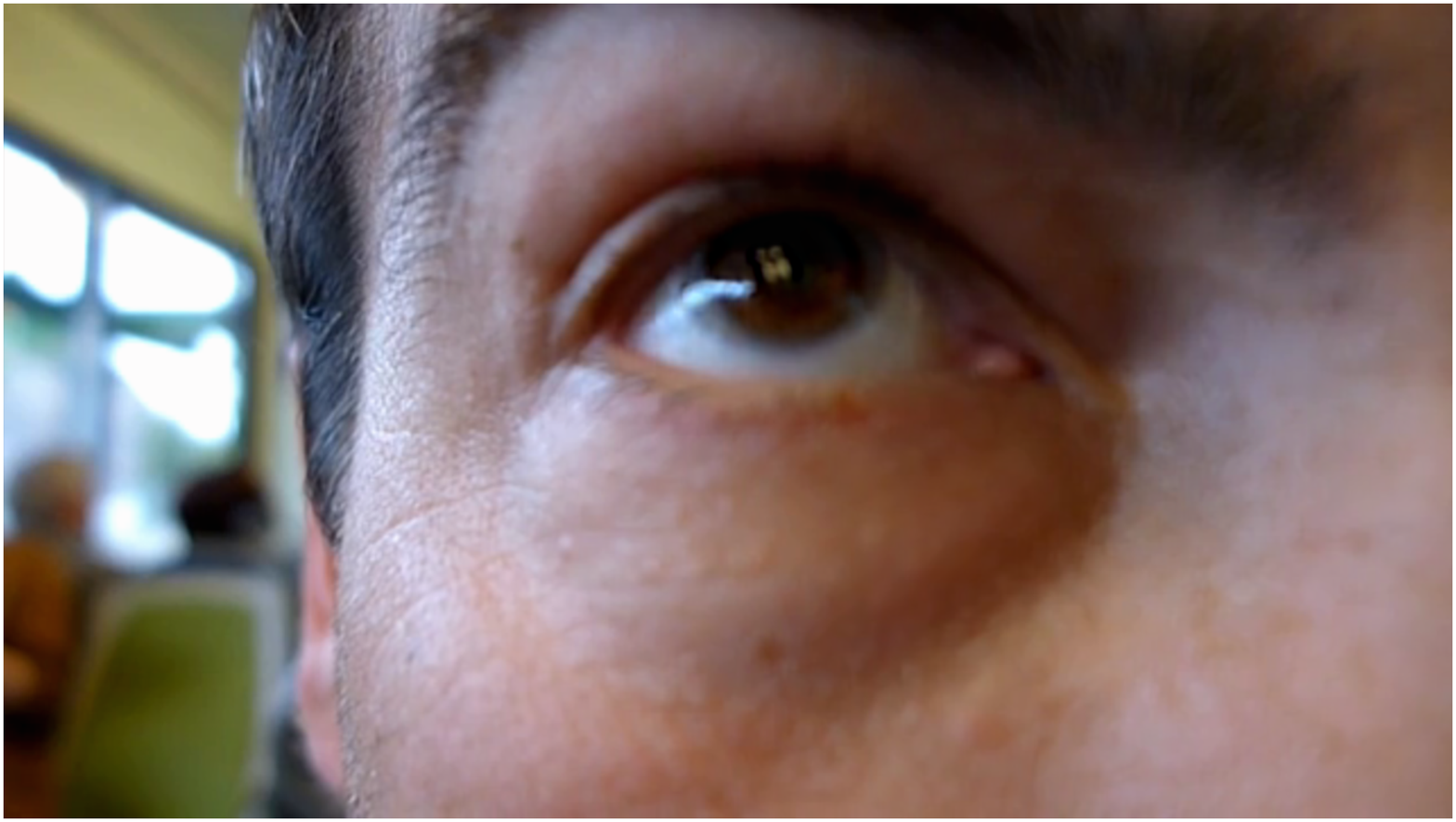}
        ~
        \includegraphics[width=0.45\textwidth]{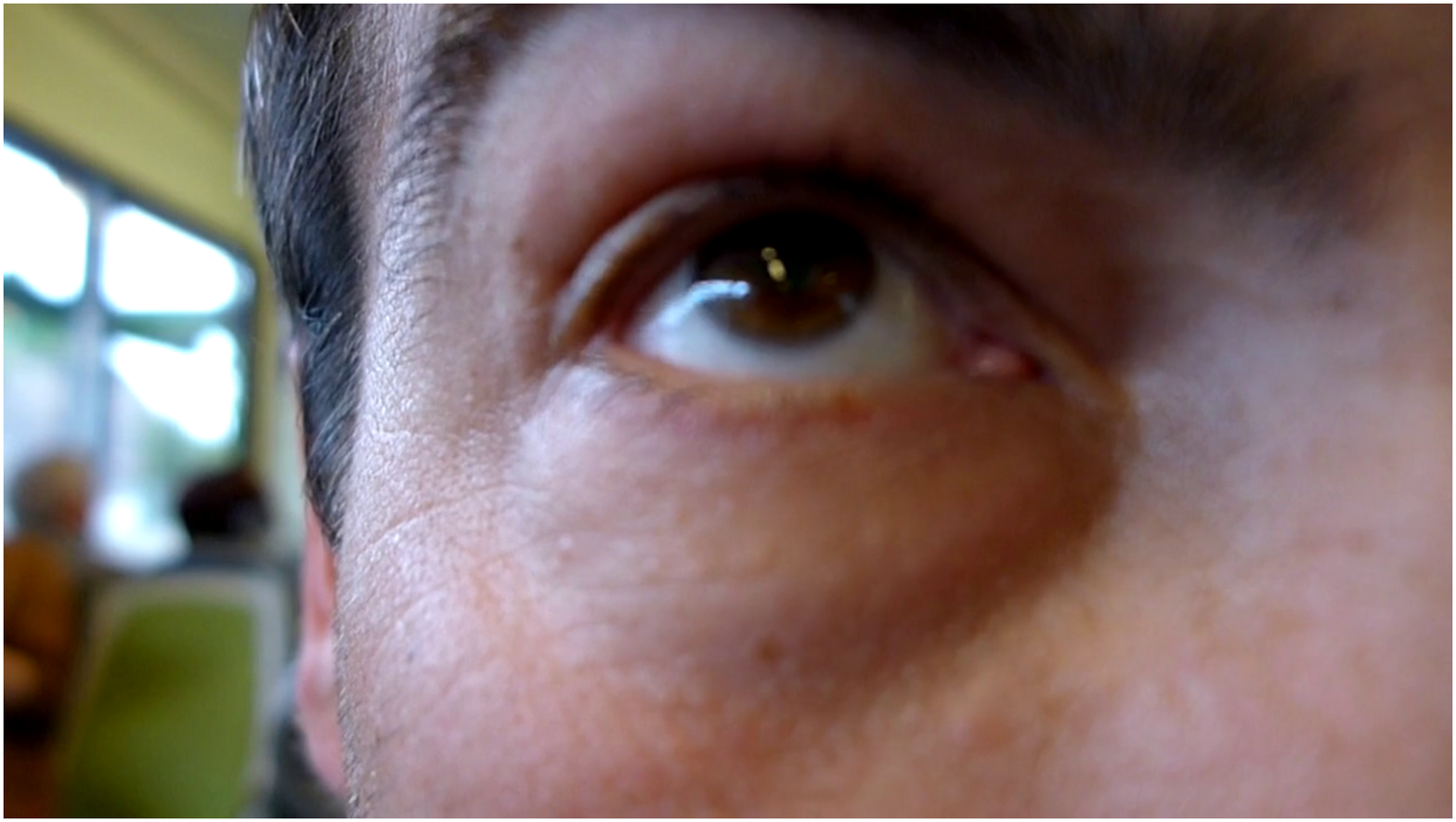}
        \caption{(Left) Example SGPLVM-generated frame and (right) HD ground truth.
        }
        \label{fig:Examples:SGPLVM:Video:Frames}
    \end{subfigure}
    
    \caption{
        (HD Video)
        (\subref{fig:Examples:SGPLVM:Video:LVs}) SGPLVM latent variables as a function of time; each curve is a separate latent dimension and markers indicate observed frames.
        (\subref{fig:Examples:SGPLVM:Video:LengthScales}) Inverse length scales learned by the SGPLVM model.
        (\subref{fig:Examples:SGPLVM:Video:Frames}) An example generated image and the corresponding ground truth.
        See the supplementary material for full-resolution pictures.
    }
    \label{fig:Examples:SGPLVM:Video}
\end{figure}

In this example, we explore the use of the SGPLVM with a dynamical prior to train on a low-resolution video sequence and impute missing frames at a higher resolution without ever showing the model any high-resolution examples.
The video we use is available at \url{https://pixabay.com/en/videos/eye-face-human-male-man-person-4376/} under the Creative Commons CC0 license and is also included in the supplementary material.
We train our model on the video with $640 \times 360$ resolution and test our predictions against the $1280 \times 720$ HD video.
Pixels in the low-resolution video are assigned spatial locations at the center of the $2 \times 2$ clusters of pixels in the high-resolution video.
The video contains $n_\xi^{tot} = 156$ frames.
We randomly select $n_\xi = 78$ frames as training data and predict on the remaining $n_\xi^* = 78$ frames.
We train a dynamical SGPLVM model with $m_\xi=n_\xi$ latent variable inducing inputs, $m_s = n_s$ spatial inducing inputs, and $d_\xi = 30$ latent dimensions.
Thus, the SGPLVM models $n=m=1.8 \times 10^7$ data. %
The spatial input structure is exploited by defining $\mtx X^{(s)} = \mtx X^{(s, 1)} \times \mtx X^{(s, 2)}$, where
$\mtx X^{(s, 1)} \in \reals^{640 \times 1}$ and
$\mtx X^{(s, 2)} \in \reals^{360 \times 1}$.
The outputs are RGB triplets; thus, $d_y = 3$.
Predictions with the dynamical SGPLVM are compared against a structured GP regression model which predicts the RGB triplets as a function of time and spatial location (3 input dimensions).
Prediction accuracy on each frame is measured via RMSE and MNLP.
For the dynamical SGPLVM, we obtain an 
RMSE of $0.13$ ($0.063$, $0.61$) and an 
MNLP of $-2.13$ ($-2.81$, $0.71$); 
for the structured GP, we obtain an 
RMSE of $0.23$ ($0.097$, $0.74$) and 
MNLP of $-0.021$ ($-0.45$, $1.06$).
Note that data are pre-processed to have zero mean and unit variance on each channel.
Thus, we find that the latent variable modeling of the frames enables us to get substantially better predictive accuracy.
Figure~\ref{fig:Examples:SGPLVM:Video} shows the time-dependent latent variable posterior means, the latent dimensions' inverse length scales, and a comparison between a frame from the HD video and the corresponding generated frame from the SGPLVM.

\section{Conclusions and Discussion}
\label{sec:Conclusion}
In this work, we derived a structured Gaussian process latent variable model extending that explicitly captures spatial correlations in high-dimensional observation data.
Computational tractability is maintained by identifying Kronecker product structure in the model kernel matrices.
The modeled spatiotemporal correlations are expressed in terms of interpretable parameterized kernel functions, allowing for one to use the generative model at a higher resolution than that of the training data.

We envision several extensions to our work.
First, the SGPLVM might be incorporated as a structured layer within a deep Gaussian process \cite{damianou2013deep} with additional hidden layers preceding or following it.
Second, we would like to explore the use of the SGPLVM as a data-driven surrogate model for modeling stochastic partial differential equations with high-dimensional nonstationary outputs, and
as a data-driven high-dimensional stochastic input model (i.e.\ for dimensionality reduction).

\subsubsection*{Acknowledgments}
This work was supported from the University of Notre Dame, the Center for Research Computing (CRC) and the Center for Informatics and Computational Science (CICS). 
Early developments were supported by the Computer Science and Mathematics Division of ORNL under the DARPA EQUiPS program. 
N.Z.\ thanks the Technische Universit\"{a}t M\"{u}nchen, Institute for Advanced Study for support through a Hans Fisher Senior Fellowship, funded by the German Excellence Initiative and the European Union Seventh Framework Programme under Grant Agreement No.\ 291763.

\bibliographystyle{elsarticle-num}

\end{document}